\documentclass[letterpaper]{article}
\usepackage{proceed2e}
\usepackage[margin=1in]{geometry}
\usepackage[round]{natbib}

\usepackage{times}
\usepackage[nolist]{acronym}
\usepackage{hyperref}
\usepackage{etoolbox}
\usepackage{multirow}
\usepackage{graphicx}
\usepackage{amsmath}
\usepackage{color}

\newcommand{\weight}{\mathbf{\theta}}

\title{Make (Nearly) Every Neural Network Better: Generating Neural Network Ensembles by Weight Parameter Resampling}

\author{ {\bf Jiayi Liu, Samarth Tripathi, Unmesh Kurup, Mohak Shah} \\
Advanced AI  \\
LG Electronics\\
Santa Clara, CA, USA 
}

\begin{document}

\begin{acronym}
\acro{DNN}{Deep Neural Network}
\acro{SGD}{Stochastic Gradient Descent}
\acro{SWA}{Stochastic Weight Averaging}
\acro{Reptu}{Resample Parameter by Training Uncertainty}
\end{acronym}
\maketitle

\begin{abstract}
\acp{DNN} have become increasingly popular in computer vision, natural language processing, and other areas. However, training and fine-tuning a deep learning model is computationally intensive and time-consuming. We propose a new method to improve the performance of nearly every model including pre-trained models. The proposed method uses an ensemble approach where the networks in the ensemble are constructed by reassigning model parameter values based on the probabilistic distribution of these parameters, calculated towards the end of the training process. For pre-trained models, this approach results in an additional training step (usually less than one epoch). We perform a variety of analysis using the MNIST dataset and validate the approach with a number of \ac{DNN} models using pre-trained models on the ImageNet dataset.
\end{abstract}

\acresetall

\section{Introduction}\label{sec:intro}
\acp{DNN} have applications in image classification, object detection, machine translation, and many others\citep{he2016deep, redmon2016you, wu2016google}.
In such applications, even a marginal improvement in model performance can have significant business value. 

Ensemble methods are commonly used in computer vision competitions and achieve better performance comparing compared to single models \citep{krizhevsky2012imagenet, Simonyan2015, he2016deep}.
However, in the case of \acp{DNN}, training even a single model is computationally intensive, making ensemble approaches less tractable.

The distribution of \ac{DNN} parameters has been studied extensively as part of Bayesian Neural Networks.
The state-of-the-art variational inference provides robustness to overfitting leading to better model performance \citep{Gal2016}.
However, the information from training updates is not fully utilized.

Recently, \cite{garipov2018loss} proposed a procedure to ensemble a \ac{DNN} model at different training stages.
The method enables a fast ensemble by reducing the number of models that need to be trained from scratch.
Furthermore, the same team improved the method by directly averaging the weights instead of using an ensemble thereby reducing the computation cost \citep{Izmailov2018}.

The above-mentioned methods all require retraining the model.
We propose a new method to use the uncertainty residing in the \ac{SGD} updates for the model ensembling and parameter averaging to improve the model prediction performance.

The key contributions of the paper include:
\begin{itemize}
\item We propose a fast and universal method to finetune a given \ac{DNN} model for better prediction performance.
\item We explore and study the factors that are critical to the proposed method using MNIST dataset \citep{lecun1998gradient}.
\item We test the approach against the state-of-the-art models, i.e. Inception-V3, MobileNet \citep{szegedy2015rethink,Howard2017}, using the ImageNet dataset \citep{deng2009imagenet}.
\end{itemize}

In this paper, we first introduce our approach in Sec.~\ref{sec:method}.
Then we carry out an extensive analysis using LeNet model on MNIST dataset and evaluate the result on a variety \acp{DNN} models on ImageNet dataset in Sec.~\ref{sec:experiment}.  Finally, we discuss the proposed methods and compare with other related works in Sec.~\ref{sec:discussion} and conclude the paper.

\section{Method}\label{sec:method}

\acp{DNN} are commonly trained by the \ac{SGD} method or its variants, where the parameters, $\weight$, are updated based on the derivative of the loss for each mini-batch of data.

\begin{equation}\label{eq:update}
\weight^{(t+1)} = \weight^{(t)} - \ell\nabla_\weight \sum_{i \in \text{batch}} L_i,
\end{equation}
where $L_i$ is the loss of a sample $i$ for given model parameters $\weight^{(t)}$ at step $t$ and the hyperparameter $\ell$ is the learning rate that controls the step size of the update.

Given the variations across batches of data, the updates are stochastic and the parameters asymptotically reach local optima.
And to reduce the convergence instability,  the learning rate that throttles the steps size of updates is either predetermined as a constant or follows a learning schedule or is updated according to the update statistics.

In this paper, we propose to use the uncertainty of the model parameters during the training updates to create a final model.
We first estimate the mean and the variance of the parameters by continuing the training with a few mini-batches after the model is trained (this {\it fine-tuning} stage may or may not share the same \ac{SGD} method used in the previous training).
Because the network size is commonly very large, we uses an online algorithm to update the mean and variance \citep{Welford1962Note}, instead of saving all intermediate values:
\begin{eqnarray}
\bar\weight^{(t+1)} &=& \bar\weight^{(t)} + \frac{\weight^{(t+1)} - \bar\weight^{(t)}}{t+1} \\
\sigma^2(\weight)^{(t+1)} &=& \frac{\sigma^2(\weight)^{(t)}t + \frac{t}{t+1}(\weight^{(t+1)} - \bar\weight^{(t)})^2}{t+1}
\end{eqnarray}
And then, we use two different approaches to resample the parameters for predictions.
\begin{itemize}
\item We reassign the value of parameters to the mean $\weight^{(t)}$ after the fine-tuning stage.
\item We assign the value parameters follows a Gaussian distribution given the mean and standard deviation during the fine-tuning stage.  We create multiple models from the resampling and make predictions by ensembling the model predictions by the average.
\end{itemize}

\section{Experiment}\label{sec:experiment}
We tested our method against two experiment sets. Using MNIST, we explored a large number of configurations to understand the limiting factors in Sec.~\ref{sec:mnist}. And we provided a number of results from pre-trained models on ImageNet to examine the robustness of the method in Sec.\ref{sec:dnn_results}.

\subsection{MNIST}\label{sec:mnist}
\begin{figure*}[htbp]
\begin{center}
\includegraphics[width=\textwidth]{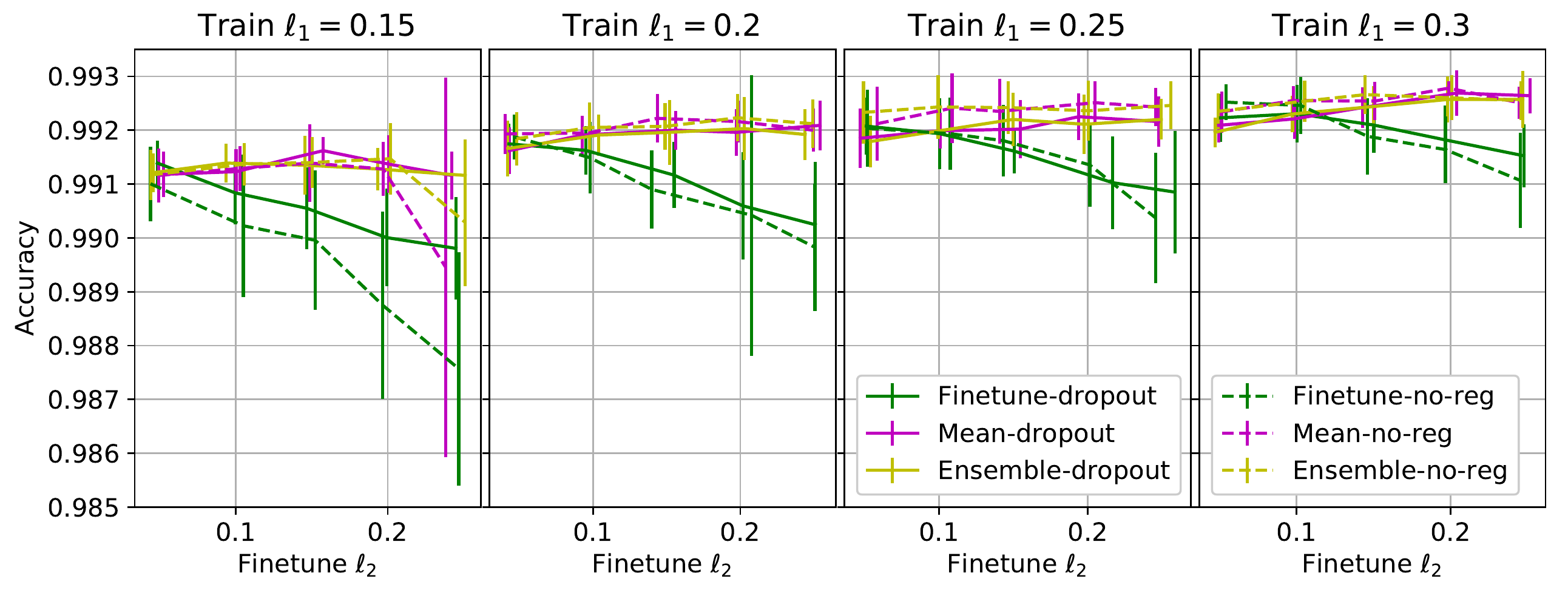}
\caption{{\bf MNIST Accuracies with Different Learning Rates, Optimized by \ac{SGD}}}
\label{fig:mnist_diff_lr}
\end{center}
\end{figure*}

\subsubsection{Setup}
We used MNIST dataset \citep{lecun1998gradient} to quickly explore the configurations of the LeNet model, namely optimization and regularization, and to understand important factors in the proposed method.

MNIST is a commonly used dataset for computer vision, which contains hand-written digits that split into 60000 training samples and 10000 testing samples.
We train our model on the full training set with a batch size of 128 and measure the accuracies on the testing set.
And for each configuration, we repeat the same procedure 10 times and report the mean and standard deviation of the accuracies.

The model improvement is sensitive to the final status of the pre-trained model.
In the extreme case, a model at the global minimum cannot be further improved without overfitting the data.
We choose different learning rates for training, $\ell_1$, to examine the proposed method.
A small $\ell_1$ leads to local minima that might be far away from the global one, while a large value prevents the model from settling into the minima.
In this paper, we trained the model using $\ell_1 \in {0.15,0.2,0.25,0.3}$ with 2 epochs.
We found a larger or smaller $\ell_1$ has deteriorated performance and we excluded them from the discussion.
Similarly, at the fine-tuning stage, the learning rate is also important and we tested with $\ell_2\in\{0.05,0.1,0.15,0.2,0.25\}$.
The weight distribution is estimated from updates from 500 mini-batches initialized from the pre-trained model (roughly one epoch).

Besides the learning rate, the optimization method for the model update is also important.
We mainly focused on using the plain \ac{SGD} method to understand the method behavior.
Many other update strategies have been proposed for better convergence rate, e.g. AdaGrad, Adam \citep{Duchi2011, Kingma2015}, which adaptively adopt the learning rate for a faster and better convergence.
In this paper, we also tested our method using Adam optimizer with default values in Tensorflow \footnote{See \url{https://www.tensorflow.org/api_docs/python/tf/train/AdamOptimizer}, version r1.8.}.

Regularization method also has an important impact on the model generalization and prediction accuracy.
A generalized model alleviates from the overfitting of the training data and improves prediction accuracy.
In this work, we tested our method against models with and without Dropout \citep{srivastava2014dropout}.
We used the solid line for with Dropout and dashed line for no regularization hereafter.

\subsubsection{Results}

First, we present the result of training the models using fixed learning rate \ac{SGD} method in Fig.~\ref{fig:mnist_diff_lr}.
The finetune learning rates, $\ell_2$, are jittered around in the figure for better visualization.
As expected, in the plain \ac{SGD} approach (green colored), the combination of a larger learning rate at training stage and a smaller learning rate at finetuning stage are always preferred for the best performance, because a larger learning rate at the training stage explored a larger space for global minimal and a smaller finetuning learning rate helps convergence.
The comparison shows that the regularization helps the model to be more general and more accurate.

In comparison with the plain finetuning, a larger $\ell_2$ is always preferred in our approaches.
And we didn't see any performance degradation in a wide range of learning rates.
Also, the regularization has less impact on the model performance as the difference between dashed lines and solid lines are marginal.

Finally, we didn't see significant differences in the mean-resampled model and ensemble approach with 3 resampled models.  
However, we do see a marginal improvement with 10 ensembles but it is typically not feasible in a real application as it takes 10 times longer. 
We could treat the mean-resampled model as a special case in this ensemble approach.
Also, we found that the method need enough updates to measure the distribution reliably (one epoch is typically sufficient).

\begin{figure}[htbp]
\begin{center}
\includegraphics[width=\columnwidth]{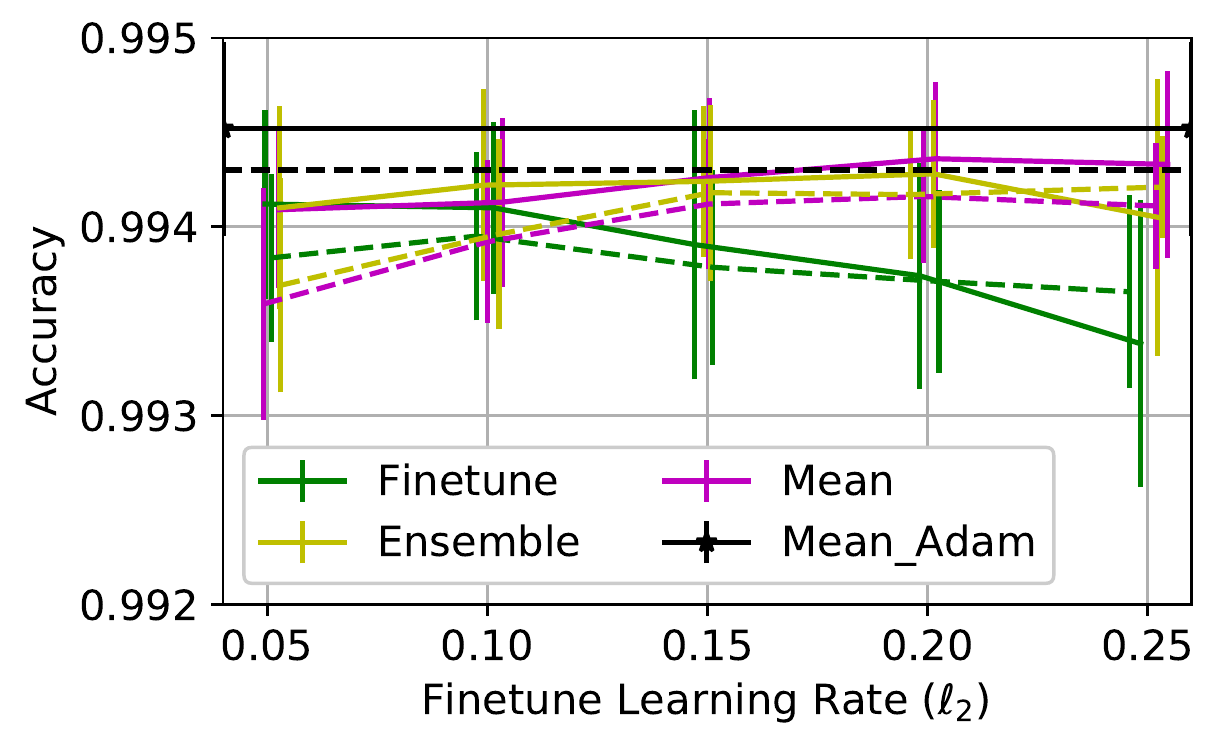}
\caption{{\bf MNIST Accuracies with Different Learning Rates, Optimized by Adam}}
\label{fig:mnist_adam}
\end{center}
\end{figure}

In Fig.~\ref{fig:mnist_adam}, we compared the result on pre-trained models that were trained using the Adam optimizer.
Again, the results from the finetune stage are similar to the \ac{SGD} results in Fig.~\ref{fig:mnist_diff_lr}.
We also performed the fine-tuning stage using the Adam optimizer with default values. 
As learning rate is not relevant for Adam, the results fine-tuned by the Adam optimizer are marked by the straight line in black.
Our resampled method gives the best result while the fine-tuned Adam result is below 0.993 (not shown).
Finally given the large scatter from the 10 different runs, we only see marginal improvement by using the regularization in our approach.

\subsection{\ac{DNN} Results}\label{sec:dnn_results}

We also performed many experiments on ImageNet \citep{deng2009imagenet} using public available pre-trained models to validate the generalization of our proposed method.
As the size of the ImageNet is large, we only used 25~\% of the full dataset in the finetuning stage to estimate the uncertainties of model parameters (10,000 updates).
Finally, the accuracies after the fine tuning are also reported, and given the computational cost, we ran only one iteration per model configuration.

\begin{figure}[h]
\centering
\includegraphics[width=\columnwidth]{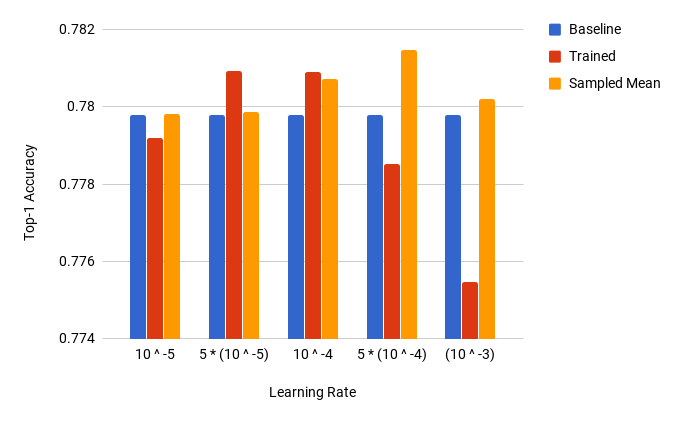}
\caption{\bf Imagenet Results Using Pre-trained Inception}
\label{fig:imagenet-inception}
\end{figure}

Fig.~\ref{fig:imagenet-inception}, we examined the pre-trained Inception-V3 model\footnote{Retrieved from \url{http://download.tensorflow.org/models/inception_v3_2016_08_28.tar.gz}} \citep{szegedy2015rethink} with a range of learning rates.
The pre-trained model is highly fine-tuned, hence the improvements are very small. 
But still, the resampled mean weights does improve upon the results from the baseline model and the best-finetuned model. 
Also, it is worth to mention that the proposed method showed a consistently better performance over a wide range of learning rates in both pre-trained models.

\begin{figure}[h]
\centering
\includegraphics[width=\columnwidth]{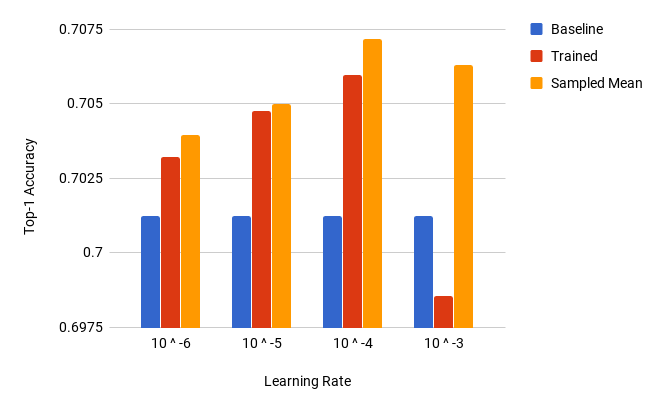}
\caption{\bf Imagenet Results Using Pre-trained MobileNet}
\label{fig:imagenet-mobilenet}
\end{figure}

Fig.~\ref{fig:imagenet-mobilenet} refers to our results from MobileNet  architecture\footnote{Retrieved from \url{https://storage.googleapis.com/mobilenet_v2/checkpoints/mobilenet_v2_1.0_224.tgz}}  \citep{Howard2017} and the pre-trained base model with a top-1 accuracy of 70.124~\% because the model is designed as a light-weight model.
We achieved some improvement over the baseline model even by just using the \ac{SGD} method to finetune the model parameters. 
And, upon resampling the weights, the results show more improvement on both models in all cases.

\section{Discussion}\label{sec:discussion}

From the previous experiment results, we justified the usability of our proposed method. In this section, we will first highlight the benefits of it and then compare it with another relevant study.

The major contributions of this work are following.
First, it is tested to improve the accuracies of a range of \ac{DNN} models.
Second, it is less sensitive to the learning rate that used to update the model parameters in the training stage.
Third, resampling the model parameters with their mean values requires no additional computing cost for the inference and a marginal burden in the training stage.
Finally, the model is efficient that it just requires one epoch or less to finetune a pre-trained \ac{DNN} model.

\cite{Izmailov2018} proposed a \ac{SWA} method to improve the model performance.
Similar to our model to use mean to reassign model parameters, it uses the average of the parameters during the training steps.
The two main differences between our approaches are:

\begin{itemize}
\item Our method is to finetune a model based on the pre-trained values, whereas \ac{SWA} method needs to train a model from scratch.  
The two methods have different focuses at the moment. 
\item We sampled the parameter distribution at each step during the fine-tuning stage, and the \ac{SWA} method samples at the end of each learning cycle.
\end{itemize}

So, we focused on the improvement of the pre-trained model rather than comparing with their approach.
And, it is interesting to compare with the \ac{SWA} method and other algorithms in future studies.

\section{Conclusion}

We concluded the paper with extensive experiments with our proposed method on the MNIST with a simple LeNet model and initial results on ImageNet data with state-of-the-art \ac{DNN} models.
Our future work includes developing a theoretical understanding of this approach that will provide the solid foundation to further guide the usability of our method.

\newpage

\bibliography{pnn}

\begin{thebibliography}{16}
\providecommand{\natexlab}[1]{#1}
\providecommand{\url}[1]{\texttt{#1}}
\expandafter\ifx\csname urlstyle\endcsname\relax
  \providecommand{\doi}[1]{doi: #1}\else
  \providecommand{\doi}{doi: \begingroup \urlstyle{rm}\Url}\fi

\bibitem[Deng et~al.(2009)Deng, Dong, Socher, Li, Li, and
  Fei-Fei]{deng2009imagenet}
Jia Deng, Wei Dong, Richard Socher, Li-Jia Li, Kai Li, and Li~Fei-Fei.
\newblock Imagenet: A large-scale hierarchical image database.
\newblock In \emph{Computer Vision and Pattern Recognition, 2009. CVPR 2009.
  IEEE Conference on}, pages 248--255. IEEE, 2009.

\bibitem[Duchi et~al.(2011)Duchi, Hazan, and Singer]{Duchi2011}
John Duchi, Elad Hazan, and Yoram Singer.
\newblock {Adaptive subgradient methods for online learning and stochastic
  optimization}.
\newblock \emph{The Journal of Machine Learning Research}, 12\penalty0
  (Jul):\penalty0 2121--2159, 2011.

\bibitem[Gal and Ghahramani(2016)]{Gal2016}
Yarin Gal and Zoubin Ghahramani.
\newblock {Dropout as a Bayesian Approximation: Representing Model Uncertainty
  in Deep Learning}.
\newblock In \emph{Proceedings of the 33 rd International Conference on Machine
  Learning}, pages 1050----1059, New York, NY, USA, 2016. JMLR.

\bibitem[Garipov et~al.(2018)Garipov, Izmailov, Podoprikhin, Vetrov, and
  Wilson]{garipov2018loss}
Timur Garipov, Pavel Izmailov, Dmitrii Podoprikhin, Dmitry Vetrov, and
  Andrew~Gordon Wilson.
\newblock {Loss Surfaces, Mode Connectivity, and Fast Ensembling of DNNs}.
\newblock Feb 2018.
\newblock \doi{arXiv:1802.10026}.

\bibitem[He et~al.(2016)He, Zhang, Ren, and Sun]{he2016deep}
Kaiming He, Xiangyu Zhang, Shaoqing Ren, and Jian Sun.
\newblock Deep residual learning for image recognition.
\newblock In \emph{Proceedings of the IEEE conference on computer vision and
  pattern recognition}, pages 770--778, 2016.

\bibitem[Howard et~al.(2017)Howard, Zhu, Chen, Kalenichenko, Wang, Weyand,
  Andreetto, and Adam]{Howard2017}
Andrew~G. Howard, Menglong Zhu, Bo~Chen, Dmitry Kalenichenko, Weijun Wang,
  Tobias Weyand, Marco Andreetto, and Hartwig Adam.
\newblock {MobileNets: Efficient Convolutional Neural Networks for Mobile
  Vision Applications}.
\newblock Apr 2017.
\newblock \doi{arXiv:1704.04861}.

\bibitem[Izmailov et~al.(2018)Izmailov, Podoprikhin, Garipov, Vetrov, and
  Wilson]{Izmailov2018}
Pavel Izmailov, Dmitrii Podoprikhin, Timur Garipov, Dmitry Vetrov, and
  Andrew~Gordon Wilson.
\newblock {Averaging Weights Leads to Wider Optima and Better Generalization}.
\newblock Mar 2018.
\newblock \doi{arXiv:1803.05407}.

\bibitem[Kingma and Ba(2015)]{Kingma2015}
Diederik~P. Kingma and Jimmy~Lei Ba.
\newblock {Adam: a Method for Stochastic Optimization}.
\newblock In \emph{International Conference on Learning Representations 2015},
  2015.

\bibitem[Krizhevsky et~al.(2012)Krizhevsky, Sutskever, and
  Hinton]{krizhevsky2012imagenet}
Alex Krizhevsky, Ilya Sutskever, and Geoffrey~E Hinton.
\newblock {ImageNet Classification with Deep Convolutional Neural Networks}.
\newblock In \emph{Advances In Neural Information Processing Systems}, pages
  1--9, 2012.
\newblock \doi{http://dx.doi.org/10.1016/j.protcy.2014.09.007}.

\bibitem[LeCun et~al.(1998)LeCun, Bottou, Bengio, and
  Haffner]{lecun1998gradient}
Yann LeCun, L{\'{e}}on Bottou, Yoshua Bengio, and Patrick Haffner.
\newblock {Gradient-based learning applied to document recognition}.
\newblock \emph{Proceedings of the IEEE}, 86\penalty0 (11):\penalty0
  2278--2324, 1998.

\bibitem[Redmon et~al.(2016)Redmon, Divvala, Girshick, and
  Farhadi]{redmon2016you}
Joseph Redmon, Santosh Divvala, Ross Girshick, and Ali Farhadi.
\newblock You only look once: Unified, real-time object detection.
\newblock In \emph{Proceedings of the IEEE conference on computer vision and
  pattern recognition}, pages 779--788, 2016.

\bibitem[Simonyan and Zisserman(2015)]{Simonyan2015}
Karen Simonyan and Andrew Zisserman.
\newblock Very deep convolutional networks for large-scale image recognition.
\newblock In \emph{International Conference on Learning Representations
  (ICRL)}, 2015.

\bibitem[Srivastava et~al.(2014)Srivastava, Hinton, Krizhevsky, Sutskever, and
  Salakhutdinov]{srivastava2014dropout}
Nitish Srivastava, Geoffrey Hinton, Alex Krizhevsky, Ilya Sutskever, and Ruslan
  Salakhutdinov.
\newblock {Dropout: A Simple Way to Prevent Neural Networks from Overfitting}.
\newblock \emph{Journal of Machine Learning Research}, 15\penalty0
  (1):\penalty0 1929--1958, 2014.

\bibitem[Szegedy et~al.(2015)Szegedy, Vanhoucke, Ioffe, Shlens, and
  Wojna]{szegedy2015rethink}
Christian Szegedy, Vincent Vanhoucke, Sergey Ioffe, Jonathon Shlens, and
  Zbigniew Wojna.
\newblock {Rethinking the Inception Architecture for Computer Vision}.
\newblock Dec 2015.
\newblock \doi{arXiv:1512.00567}.

\bibitem[Welford(1962)]{Welford1962Note}
B.~P. Welford.
\newblock Note on a method for calculating corrected sums of squares and
  products.
\newblock \emph{Technometrics}, 4\penalty0 (3):\penalty0 419--420, 1962.
\newblock ISSN 00401706.

\bibitem[Wu et~al.(2016)Wu, Schuster, Chen, Le, Norouzi, Macherey, Krikun, Cao,
  Gao, Macherey, Klingner, Shah, Johnson, Liu, Łukasz Kaiser, Gouws, Kato,
  Kudo, Kazawa, Stevens, Kurian, Patil, Wang, Young, Smith, Riesa, Rudnick,
  Vinyals, Corrado, Hughes, and Dean]{wu2016google}
Yonghui Wu, Mike Schuster, Zhifeng Chen, Quoc~V. Le, Mohammad Norouzi, Wolfgang
  Macherey, Maxim Krikun, Yuan Cao, Qin Gao, Klaus Macherey, Jeff Klingner,
  Apurva Shah, Melvin Johnson, Xiaobing Liu, Łukasz Kaiser, Stephan Gouws,
  Yoshikiyo Kato, Taku Kudo, Hideto Kazawa, Keith Stevens, George Kurian,
  Nishant Patil, Wei Wang, Cliff Young, Jason Smith, Jason Riesa, Alex Rudnick,
  Oriol Vinyals, Greg Corrado, Macduff Hughes, and Jeffrey Dean.
\newblock Google's neural machine translation system: Bridging the gap between
  human and machine translation.
\newblock Sep 2016.
\newblock \doi{arXiv:1609.08144}.

\end{thebibliography}
\bibliographystyle{plainnat}

\end{document}